\documentclass{article}
\usepackage{longtable}
\usepackage{array} 
\usepackage{float}
\usepackage{amsmath}
\usepackage{PRIMEarxiv}
\usepackage{makecell}
\usepackage[utf8]{inputenc} 
\usepackage[T1]{fontenc}    
\usepackage{hyperref}       
\usepackage{url}            
\usepackage{booktabs}       
\usepackage{amsfonts}       
\usepackage{nicefrac}       
\usepackage{microtype}      
\usepackage{lipsum}
\usepackage{fancyhdr}       
\usepackage{graphicx}       
\graphicspath{{media/}} 
\usepackage{xcolor}
\usepackage{authblk}
\usepackage{hyperref}
\usepackage{multirow} 
\usepackage{longtable} 
\usepackage{enumitem}
\usepackage{multirow}
\usepackage{array}
\usepackage{verbatim}
\begin{document}

\title{CardioBench: A Real-World Data Benchmark for Evaluating Large Language Models in Clinically Authentic Cardiovascular Care Scenarios}
\author[1,2,3,†]{Xiao Li}
\author[4,†]{Mouxiao Bian}
\author[5]{Zhaodi Wu}
\author[4]{Sijie Ren}
\author[1,2,3]{Juechen Chen}
\author[4]{Lu Lu}
\author[4]{Jingru Ding}
\author[4]{Yun Zhong}
\author[4,*]{Jie Xu}
\author[1,2,3,*]{Yixiu Liang}
\author[1,2,3]{Junbo Ge}
\affil[1]{\textit{
   Department of Cardiology, Zhongshan Hospital Fudan University  \\
    Shanhai, China
}}
\affil[2]{\textit{
    Shanghai Institute of Cardiovascular Diseases, National Clinical \\
    Shanghai, China
}}
\affil[3]{\textit{
    Research Center for Interventional Medicine  \\
    Shanghai, China
}}
\affil[4]{\textit{
    Shanghai Artificial Intelligence Laboratory\\
    Shanghai, China
}}
 \affil[5]{\textit{
   Minhang Hospital, Fudan University \\
    Shanghai, China
}}

\maketitle

\footnotetext[1]{†These authors contributed equally.}
\footnotetext[2]{*Correspondence: 
Yixiu Liang(liangyixiu@fudan.edu.cn),
Jie Xu (xujie@pjlab.org.cn)
}

\begin{abstract}
\textbf{Background: }Most medical large language model (LLM) benchmarks emphasize examination knowledge or isolated tasks and may not reflect the longitudinal, multimodal, safety-critical workflow of cardiovascular care.

\textbf{Objective: }To develop CardioBench, a real-world cardiovascular benchmark spanning the full care pathway, and to characterize model performance across clinical dimensions, specialist tasks, and complementary evaluation metrics.

\textbf{Methods: }CardioBench comprised 2,263 items across 13 task-specific datasets derived from de-identified real-world cardiovascular records and examination data. Dataset annotation, reference-answer construction, and atomic key-point definition were performed by 16 cardiology physicians holding intermediate or higher professional titles. Two cardiologists with senior professional titles then conducted cross-review and adjudication to verify clinical validity, task clarity, guideline concordance, and consistency of the reference standards. 7 LLMs generated 15,841 outputs under standardized zero-shot inference settings. Open-ended tasks were evaluated using complementary key-point coverage and holistic clinical-quality components, whereas the five-option clinical ethics task was scored by accuracy.

\textbf{Results: }GPT-5.4 achieved the highest overall macro-average (62.55) and item-weighted mean (62.19), followed by Gemini 3.1 Pro (59.95) and Qwen 3.6 27B (59.72). GPT-5.4 ranked first in all three dimensions, scoring 70.42 in full-cycle care, 58.47 in multimodal interpretation and documentation, and 43.17 in communication, safety, and ethics. CardioAuxReport had the highest cross-model mean (86.38), followed by CardioChronicMed (75.04), CardioAdmRec (74.66), and CardioChronicHealth (70.16). CardioECGRead (17.25) and CardioEthics (17.34) were the weakest tasks. GPT-5.4 led six tasks, Qwen 3.6 27B led four, Grok 4.3 led two, and Gemini 3.5 Flash led one. The largest differences between holistic clinical judgment and key-point coverage occurred in CardioComm (52.71), CardioEmergRescue (52.05), and CardioTreatPlan (48.80).

\textbf{Conclusions: }To our knowledge, CardioBench is the largest real-world, multi-task benchmark developed specifically for LLM evaluation across the cardiovascular care continuum and provides the broadest coverage of clinically authentic cardiology scenarios reported to date. Its specialist-led annotation and senior cross-review framework provides a rigorous foundation for identifying task-specific strengths, clinically important omissions, and priorities for future model development.

\end{abstract}


\section{Introduction}
Cardiovascular disease remains the leading cause of death and disability worldwide and creates a sustained need for rapid diagnosis, quantitative risk assessment, acute stabilization, secondary prevention, and long-term follow-up\cite{mensah2023global,martin20252025}. The clinical burden is not concentrated in a single decision point. Instead, cardiovascular care unfolds across repeated encounters in which clinicians must reconcile symptoms, comorbidities, medications, serial laboratory values, physiological measurements, imaging, and patient preferences. This longitudinal and multimodal structure makes cardiovascular medicine a particularly demanding setting for evaluating general-purpose artificial intelligence.

Large language models (LLMs) have demonstrated broad medical knowledge, strong performance on licensing-style examinations, and increasingly capable clinical text generation\cite{singhal2022large,singhal2025toward,van2024adapted,mcduff2025towards,tu2025towards}. However, systematic evaluations have cautioned that high scores on structured medical tasks may not translate directly into dependable clinical decision-making\cite{hager2024evaluation,bedi2025testing,agrawal2025evaluation}. Recent studies have also moved beyond static question answering toward real-world case evaluation, dynamic diagnostic procedures, clinical prediction, and competency-based benchmarks\cite{sandmann2025benchmark,chen2026grounding,zhu2026clinicrealm,li2026evaluating}. These studies collectively show that strong performance on well-structured questions does not guarantee reliable execution of a complete clinical workflow. Performance may change when information is incomplete, temporally distributed, image based, operationally constrained, or safety critical.

Cardiovascular workflows combine several reasoning modes that are rarely assessed together. Documentation tasks require faithful extraction and chronological synthesis. Diagnostic and treatment tasks require integration of symptoms, risk factors, test findings, contraindications, and guideline recommendations. Emergency care requires recognition of subtle deterioration before the diagnosis is explicitly stated, followed by prioritized actions under time pressure. Chronic disease management requires continuity across visits, medication adjustment, monitoring for adverse effects, and individualized lifestyle guidance. Communication and ethics add another layer by requiring explanations of uncertainty, informed consent, shared decision-making, capacity assessment, privacy protection, and end-of-life considerations.

Multimodal interpretation is a central source of difficulty. Twelve-lead ECGs and cardiovascular images encode spatial and temporal information that cannot be reduced to ordinary prose. Rhythm diagnosis, interval measurement, morphology, ischemic change, lesion localization, plaque characterization, and stenosis grading require domain-specific visual reasoning and calibrated uncertainty. Earlier evaluations showed that vision-enabled LLMs could answer selected medical-image questions while remaining unreliable for open-ended interpretation\cite{jin2024hidden,zhu2024multimodal,engelstein2025effectiveness,soubh2026performance}. A 2025 comparison of multimodal LLMs with a dedicated ECG system further showed substantially lower diagnostic performance for general-purpose models, reinforcing the need to distinguish broad language competence from validated signal interpretation\cite{lee2025comparative}.

Cardiology-specific evidence remains heterogeneous and comparatively narrow in scope. Existing studies have emphasized examination-style cardiology questions\cite{skalidis2023chatgpt}, broad reviews of potential applications\cite{sarraju2023opportunities,boonstra2024artificial}, retrieval-augmented responses to acute coronary syndrome guidance\cite{alexandrou2025performance}, assistance in complex genetic cardiomyopathy management\cite{o2026large}, or standardized cardiovascular vignettes\cite{zhang2025benchmarking}. These studies provide important evidence for individual applications but do not jointly evaluate documentation, diagnostic reasoning, risk calculation, treatment planning, emergency recognition and rescue, multimodal interpretation, longitudinal management, communication, and ethics within one specialty-specific framework.

The evaluation of open-ended clinical responses presents a parallel measurement problem. Exact answer matching is unsuitable because clinically acceptable responses may differ in phrasing and structure. Key-point coverage provides an auditable measure of whether required diagnoses, actions, contraindications, monitoring steps, and escalation criteria were included. Holistic clinical evaluation can recognize semantic equivalence, organization, and overall plausibility, but may be influenced by fluency, verbosity, evaluator preference, and rubric design\cite{croxford2025evaluating,zhou2025automating,zheng2023judging,johri2025evaluation,tam2024framework}. A rigorous benchmark should therefore preserve complementary views of response quality rather than compressing all performance into a single opaque indicator.

We developed CardioBench to address this gap. It contains 2,263 items across 13 task-specific datasets and three high-level dimensions: full-cycle cardiovascular decision-making and longitudinal care; multimodal interpretation and clinical documentation; and communication, humanistic care, safety, and ethics. To our knowledge, CardioBench is the largest real-world, multi-task benchmark specifically developed for evaluating LLMs across the cardiovascular care continuum and offers the broadest coverage of clinically authentic cardiology scenarios reported to date. Its distinctiveness lies in combining heterogeneous clinical inputs, longitudinal information, multimodal interpretation, safety-critical decisions, communication, and ethics within a single evaluation framework. The benchmark was annotated by 16 cardiology physicians holding intermediate or higher professional titles and cross-reviewed by two cardiologists with senior professional titles, providing a rigorous basis for objective, consistent, and stable evaluation.
\section{Method}
\subsection{Study design}

We conducted a retrospective benchmark-development and comparative model-evaluation study using de-identified real-world cardiovascular data. The unit of analysis was a task-specific clinical item that reflected a plausible activity performed in cardiovascular practice, including record synthesis, diagnosis and differential diagnosis, named risk scoring, treatment planning, early recognition of deterioration, emergency management, ECG or cardiovascular image interpretation, longitudinal health and medication management, doctor-patient communication, and clinical ethics(Table \ref{tab:dataset}). CardioBench was intended for comparative capability assessment, error analysis, and research prioritization rather than autonomous clinical decision-making.

The benchmark was organized to reproduce the sequence of cardiovascular care rather than a collection of unrelated examination questions. The design therefore retained clinically relevant chronology, comorbidity, medication exposure, repeated measurements, competing diagnoses, and distractor information. Recent benchmark research has similarly emphasized real-world cases, dynamic procedures, competency-based evaluation, and expert annotation as necessary complements to examination-style testing\cite{sandmann2025benchmark,chen2026grounding,zhu2026clinicrealm,li2026evaluating}.
\begin{longtable}{| >{\raggedright\arraybackslash}p{0.18\linewidth}
                 | >{\raggedright\arraybackslash}p{0.12\linewidth}
                 | >{\raggedright\arraybackslash}p{0.20\linewidth}
                 | >{\raggedright\arraybackslash}p{0.25\linewidth}
                 | >{\raggedright\arraybackslash}p{0.11\linewidth} |}
\caption{Clinical scope and task characteristics of the 13 CardioBench datasets}
\label{tab:dataset}\\
\hline
\textbf{Dataset} & \textbf{Dimension} & \textbf{Clinical content / input} & \textbf{Expected output / capability assessed} & \textbf{Format} \\
\hline
\endfirsthead
\caption*{Table \ref{tab:dataset} (Continued)}\\
\hline
\textbf{Dataset} & \textbf{Dimension} & \textbf{Clinical content / input} & \textbf{Expected output / capability assessed} & \textbf{Format} \\
\hline
\endhead
\hline
\endfoot
\hline
\endlastfoot

\textbf{CardioAdmRec}
& Multimodal \& documentation
& Outpatient history, pre-admission laboratory and examination results, initial physical findings, and other admission-related clinical information.
& A structured cardiology admission record with source fidelity, coherent chronology, appropriate terminology, and clinically usable organization.
& Open response \\
\hline
\textbf{CardioAuxReport}
& Multimodal \& documentation
& Multiple laboratory, imaging, ultrasound, and other auxiliary examination reports recorded at different time points.
& A chronologically organized auxiliary-examination section that accurately integrates multi-source results into clinical-documentation format.
& Open response \\
\hline
\textbf{CardioDiagDiff}
& Full-cycle care
& Real-world inpatient histories, physical examinations, laboratory results, imaging, and clinically relevant temporal information.
& The principal diagnosis, clinically relevant differential diagnoses, and supporting evidence, including recognition of urgent or high-risk findings.
& Open response \\
\hline
\textbf{CardioRiskScore}
& Full-cycle care
& Case information required by a named cardiovascular risk instrument, embedded within a detailed vignette containing relevant and distracting variables.
& The requested numerical score, risk category, and clinically appropriate interpretation using the specified scoring system.
& Open response \\
\hline
\textbf{CardioTreatPlan}
& Full-cycle care
& Diagnoses, symptoms, examination findings, laboratory and imaging results, current therapies, comorbidities, and recent clinical changes.
& An individualized diagnostic and treatment plan covering priorities, medication or procedural decisions, contraindications, monitoring, and follow-up.
& Open response \\
\hline
\textbf{CardioEmergAlert}
& Full-cycle care
& Serial clinical observations and evolving test results before a cardiovascular emergency has been explicitly declared.
& Recognition of early deterioration, identification of the likely emergency, assessment of urgency, and recommendation for escalation or immediate evaluation.
& Open response \\
\hline
\textbf{CardioEmergRescue}
& Full-cycle care
& An established acute or critical cardiovascular presentation with vital signs, investigations, comorbidities, and treatment constraints.
& Prioritized emergency management, including stabilization, definitive treatment, monitoring, contraindications, and escalation thresholds.
& Open response \\
\hline
\textbf{CardioECGRead}
& Multimodal \& documentation
& ECG images input, with limited clinical context.
& Identification of rhythm and major electrocardiographic abnormalities, with a concise diagnostic interpretation supported by the displayed tracing.
& Image + open response \\
\hline
\textbf{CardioImgRead}
& Multimodal \& documentation
& Cardiovascular imaging inputs, principally coronary CTA images or structured imaging information, with limited clinical context.
& Accurate and structured description of cardiovascular imaging findings and an appropriately bounded clinical interpretation.
& Image + open response \\
\hline
\textbf{CardioChronicHealth}
& Full-cycle care
& Longitudinal follow-up data, including blood pressure, lipid levels, cardiac function, symptom changes, lifestyle factors, and adherence information.
& Individualized nonpharmacological management covering diet, exercise, weight, smoking, self-monitoring, risk-factor control, and follow-up.
& Open response \\
\hline
\textbf{CardioChronicMed}
& Full-cycle care
& Longitudinal diagnoses, medication records, treatment response, laboratory monitoring, adverse-effect risks, and changes in clinical status.
& An individualized medication-management plan with continuation, initiation, dose adjustment or discontinuation decisions, safety monitoring, and follow-up.
& Open response \\
\hline
\textbf{CardioComm}
& Communication, safety \& ethics
& Realistic cardiology communication scenarios involving condition explanation, informed consent, medication questions, risk disclosure, or adverse-event communication.
& A physician-style response integrating factual explanation, risk communication, empathy, shared decision-making, and an appropriate follow-up plan.
& Open response \\
\hline
\textbf{CardioEthics}
& Communication, safety \& ethics
& Five-option cardiovascular clinical-ethics scenarios involving autonomy, consent, confidentiality, proportionality, professional duties, and competing clinical interests.
& Selection of the single best ethically and clinically appropriate action.
& Five-option single-choice \\
\end{longtable}

\subsection{Real-world data sources and preprocessing}

The source materials consisted of de-identified cardiovascular clinical records and examination data supplied by Zhongshan Hospital, Fudan University, as described in the benchmark specification. The source pool covered inpatient and outpatient encounters and included histories, physical examinations, laboratory results, medication records, ECGs, coronary CTA and other cardiovascular imaging reports, serial follow-up data, acute deterioration episodes, and communication or ethics scenarios.

A source episode was eligible when the available information was sufficient to support at least one prespecified task without requiring the model to invent missing patient facts. Cases were excluded when the target output depended on unavailable information, the temporal sequence could not be reconstructed, the record contained unresolved internal contradictions that prevented a defensible reference, the image quality was inadequate for the intended interpretation, or a substantially duplicate scenario had already been retained. Patient identifiers and direct institutional identifiers were removed before benchmark preparation. Dates and narrative fields were normalized only when required for readability and temporal ordering; clinically meaningful values, medication histories, comorbidities, and distractors were retained.

Where one clinical episode supported more than one task, each task instance was formulated independently so that the requested output and reference standard were specific to that task. The same source information was not automatically reused across all task types. This approach reduced artificial cueing and allowed the benchmark to evaluate distinct clinical competencies rather than repeated paraphrases of one case. The final dataset contained 2,263 items, including 2,048 open-ended items and 215 five-option ethics questions.

\subsection{Task-specific scenario construction}

Benchmark preparation followed a staged workflow. First, clinically relevant source elements were extracted and organized into a chronological case representation. Second, the information required for the target task was selected while preserving realistic context, comorbidities, medication exposure, and nondecisive findings. Third, a task-specific prompt was written to request a directly usable clinical output. Fourth, a structured reference answer and a set of atomic clinical key points were prepared for open-ended tasks. Each key point was intended to represent one independently assessable fact, decision, safety requirement, monitoring action, or escalation criterion.

Documentation tasks preserved the source chronology and required clinically usable structure without unsupported additions. Diagnostic and treatment tasks included sufficient positive and negative evidence to support differential reasoning. Risk-scoring tasks specified the named instrument and version while retaining irrelevant clinical details to test variable identification. Early-warning cases presented evolving information before the emergency diagnosis was explicitly stated, whereas rescue cases required prioritized management after the critical condition had been established. Multimodal items paired prepared ECG or cardiovascular images with only the clinical context intended for the model. Communication cases required a physician-style response, and ethics questions used five plausible options with one best answer.

\subsection{Expert annotation and quality control}

Dataset annotation and reference construction were performed by 16 cardiology physicians holding intermediate or higher professional titles. The annotators transformed eligible clinical scenarios into standardized benchmark items, specified the expected response format, constructed clinically sufficient reference answers, and defined atomic key points for the open-ended tasks. For CardioEthics, the annotation process additionally included selection of the single best answer and construction of clinically plausible distractors. All items subsequently underwent cross-review by two cardiologists holding senior professional titles. The senior reviewers assessed clinical validity, source fidelity, task relevance, answerability, guideline concordance, ambiguity, and consistency between each scenario and its reference standard. Items were revised or excluded when they contained unsupported conclusions, insufficient clinical context, duplicated or highly similar scenarios, multiple equally defensible answers, inconsistent answer keys, or reference content beyond the scenario. Disagreements were resolved through discussion and consensus, with the relevant annotators consulted when clarification was required. This separation of primary annotation from senior cross-review reduced reliance on individual judgment and strengthened the objectivity, consistency, and stability of the benchmark.

Reference answers were designed to represent a clinically sufficient response rather than an exhaustive textbook discussion. Documentation references emphasized fidelity, chronology, cardiovascular terminology, and usability. Decision and emergency references specified diagnoses, supporting evidence, action priorities, monitoring, contraindications, and escalation thresholds. Multimodal references captured the target rhythm or imaging findings at the level supported by the input. Longitudinal references included individualized treatment or lifestyle changes and safety monitoring. Communication references combined factual explanation, risk communication, empathy, shared decision-making, and follow-up. Ethics items were reviewed for a single best answer, option plausibility, and consistency between the keyed answer and the scenario.

Named risk-score references were aligned with original or guideline-endorsed definitions of GRACE, CHA2DS2-VASc, and HAS-BLED\cite{granger2003predictors,eagle2004validated,lip2010refining,pisters2010novel}. Acute coronary syndrome, heart failure, atrial fibrillation, and coronary CTA items were aligned with contemporary clinical guidelines and reporting standards\cite{rao20252025,mcdonagh20232023,van20242024,cury2022cad}. The ethics domain was informed by cardiovascular professionalism and ethics guidance\cite{benjamin20212020}.

\subsection{Evaluation metrics and scoring framework}

The 12 open-ended datasets were evaluated using two complementary components. Key-point macro-recall measured recovery of prespecified clinical content. For each item, the number of matched atomic key points was divided by the total number of reference key points, and item-level recall was averaged within each LLM-task combination. A holistic clinical-quality score assessed the response against the reference and task-specific rubric, including clinical correctness, completeness, internal consistency, professional organization, and safety requirements.

A prespecified composite score combined the two open-ended components, with greater emphasis on explicit recovery of clinically required content while retaining assessment of global clinical quality. Both component scores and the composite score were reported on a 0-100 scale. CardioEthics was scored as the percentage of correctly answered questions. Because global judgments may be influenced by presentation and evaluator preference, the two open-ended components were also retained separately and compared by task\cite{croxford2025evaluating,zhou2025automating,zheng2023judging}.

\subsection{Evaluated LLMs and inference configuration}

7 LLMs were evaluated exactly as recorded in the updated results workbook: Gemini 3.5 Flash, Qwen 3.6 27B, Grok 4.3, Gemini 3.1 Pro, GPT-5.4, GPT-5.4 Mini, and Qwen 3.5 397B-A17B(Table \ref{tab:Introduction of LLMs}). Each LLM was evaluated on every benchmark item using the same task-specific prompt template and the same clinical input for that item. The evaluation was zero-shot and single-turn. No retrieval system, external clinical tool, calculator, web search, or conversation memory was available to the tested LLMs. Prompts requested the final clinical output and did not request disclosure of chain-of-thought reasoning.

Inference used one completion per item with deterministic decoding. Temperature was set to 0.0, and top-p was set to 1.0 where the provider exposed that parameter; frequency and presence penalties were left at 0 where supported. Maximum output length was selected by task type to permit a complete response without encouraging unnecessary verbosity. Multimodal inputs were submitted in the prepared benchmark image format without diagnostic overlays or derived labels. For transport errors, timeouts, or invalid structured responses, the identical request was retried up to five times, and the first valid response was retained.

\begin{table}
\centering
\caption{Evaluated large language models in CardioBench}
\label{tab:Introduction of LLMs}
\begin{tabular}{>{\raggedright\arraybackslash}p{0.12\linewidth}>{\raggedright\arraybackslash}p{0.15\linewidth}>{\raggedright\arraybackslash}p{0.12\linewidth}>{\raggedright\arraybackslash}p{0.2\linewidth}>{\raggedright\arraybackslash}p{0.09\linewidth}>{\raggedright\arraybackslash}p{0.12\linewidth}}
\toprule
\textbf{\textbf{Model}} & \textbf{\textbf{Organization}} & \textbf{\textbf{Model family}} & \textbf{\textbf{Modality type}} & \textbf{\textbf{Open source/open weight}} & \textbf{\textbf{Release date}} \\
\midrule
GPT-5.4 & OpenAI & GPT-5 series & Multimodal, text and image input & No & March 2026 \\

GPT-5.4 Mini & OpenAI & GPT-5 series & Multimodal, text and image input & No & March 2026 \\

Gemini 3.1 Pro & Google DeepMind & Gemini 3 series & Multimodal, including text and image input & No & February 2026 \\

Gemini 3.5 Flash & Google DeepMind & Gemini 3 series & Multimodal, including text and image input & No & May 2026 \\

Grok 4.3 & xAI & Grok 4 series & Multimodal, text and image input & No & May 2026 \\

Qwen 3.6 27B & Alibaba Cloud, Qwen Team & Qwen 3.6 series & Multimodal, text and image input & Yes & April 2026 \\

Qwen 3.5 397B-A17B & Alibaba Cloud, Qwen Team & Qwen 3.5 series & Multimodal, text and image input & Yes & April 2026 \\
\bottomrule

\end{tabular}

\end{table}

\subsection{Statistical analysis and visualization}

For each LLM, we calculated the mean of the tasks within each high-level dimension, an unweighted macro-average across all 13 tasks, and an item-weighted mean using the number of questions in each task. At task level, we summarized the cross-LLM mean, minimum, maximum, standard deviation, range, leading LLM, runner-up, and winning margin. For the open-ended tasks, we calculated the task-mean holistic clinical-quality score, task-mean key-point macro-recall, and their difference. Task leadership counts and rank changes across dimensions were also summarized.

Analyses were descriptive because the available workbook contained aggregate LLM-task results rather than item-level predictions. Consequently, item-level confidence intervals, paired significance tests, calibration analyses, subgroup analyses, and patient-level clustering were not estimated. All calculations used unrounded values, and reported scores were rounded to two decimal places. Analyses and figures were generated in Python using the accompanying reproducible script.
\section{Result}
\subsection{Benchmark composition}

\begin{figure}
    \centering
    \includegraphics[width=1\linewidth]{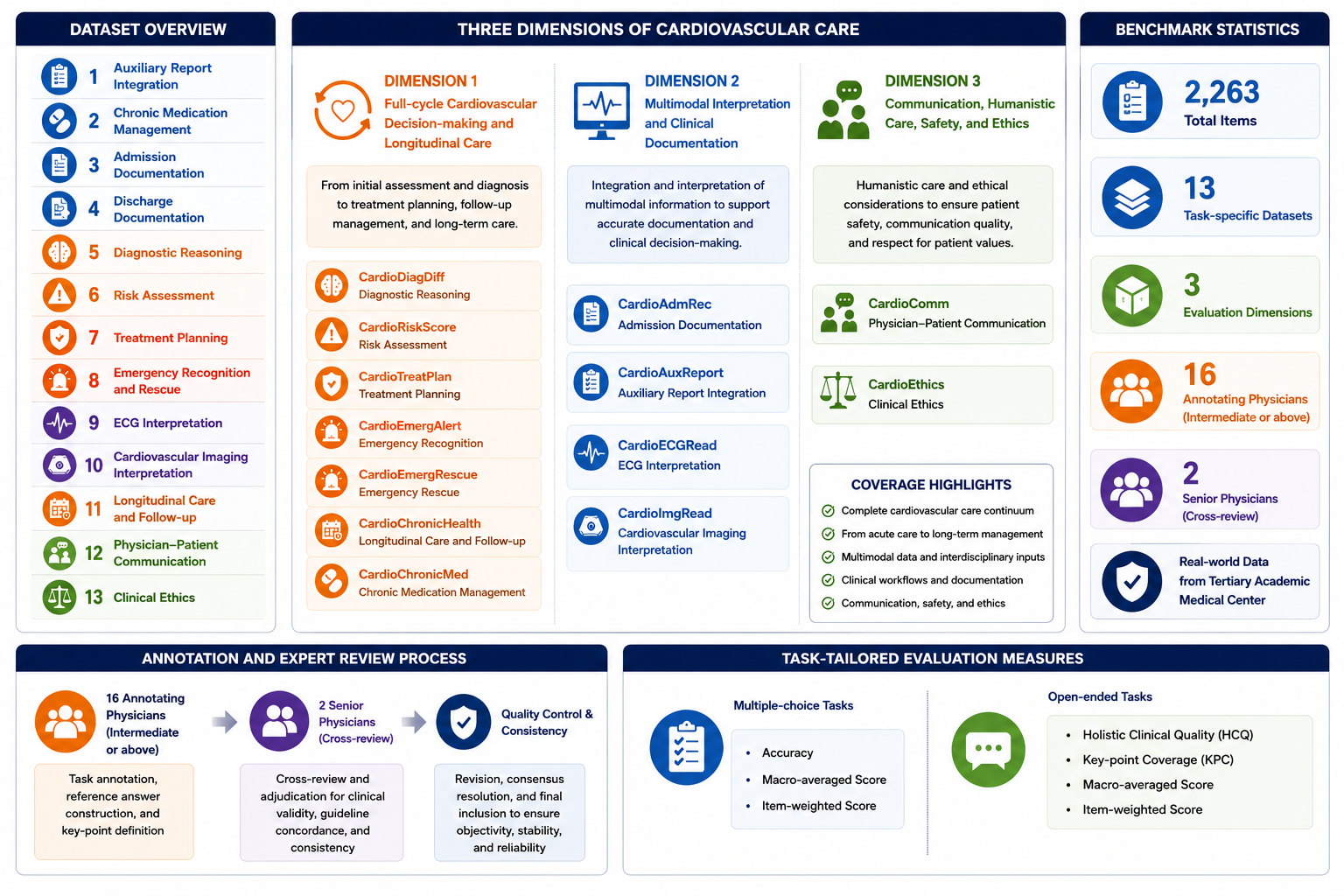}
    \caption{Overview of the CardioBench construction and evaluation framework}
    \label{fig:figure1}
\end{figure}

CardioBench included 2,263 items: 2,048 open-ended items and 215 five-option ethics questions. The 7 LLMs generated 15,841 responses. Task size ranged from 149 items for CardioRiskScore to 215 items for CardioEthics. The benchmark covered documentation, diagnostic reasoning, risk scoring, treatment planning, emergency recognition and management, multimodal interpretation, longitudinal management, communication, and ethics(Figure \ref{fig:figure1}).
\subsection{Overall performance across clinical dimensions}

\begin{figure}
    \centering
    \includegraphics[width=1\linewidth]{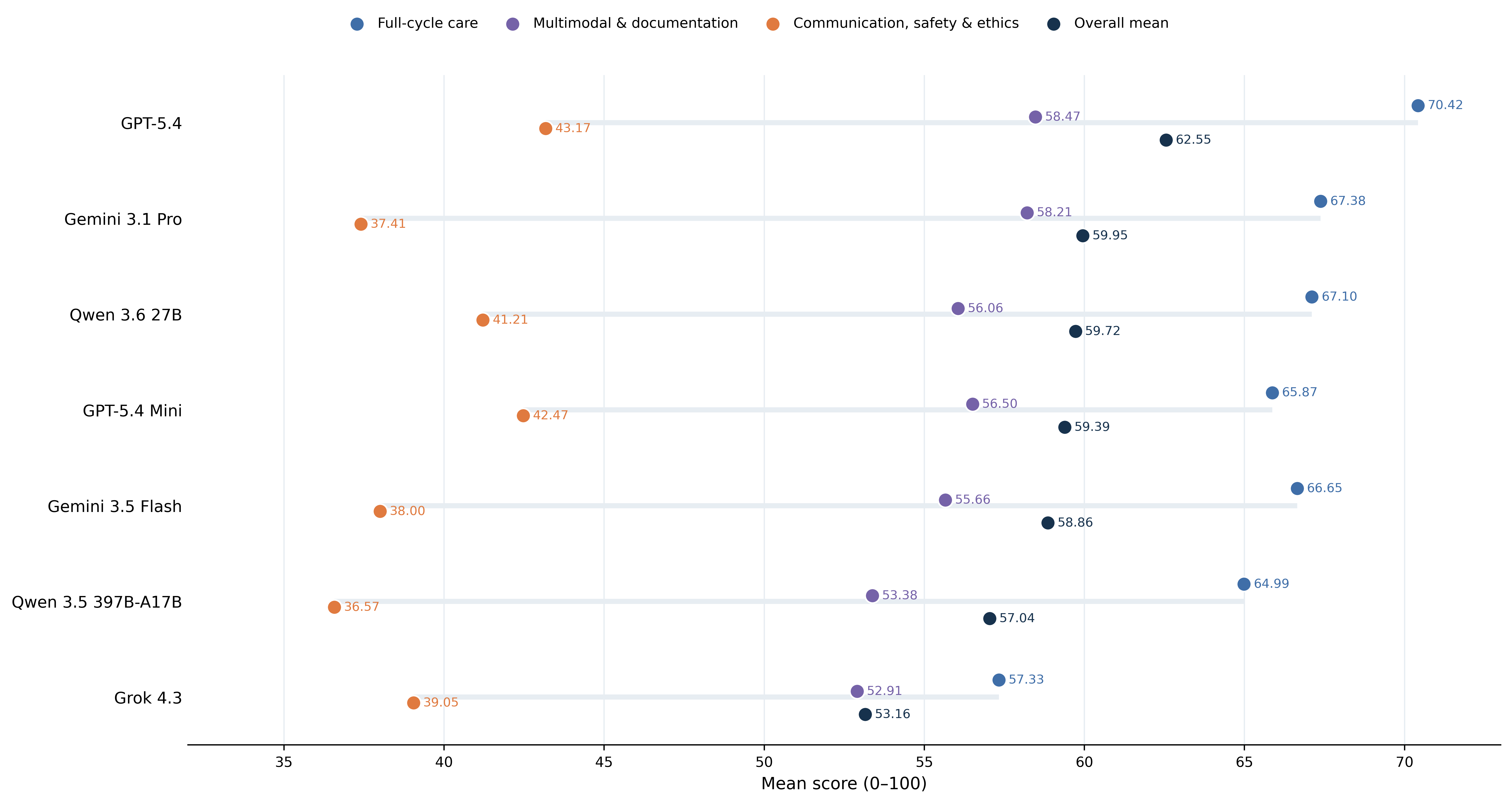}
    \caption{Dimension-specific and overall model performance}
    \label{fig:Figure2}
\end{figure}

Model performance differed across the three major clinical dimensions (Figure \ref{fig:Figure2}). GPT-5.4 achieved the highest overall macro-average across the 13 tasks at 62.55 and the highest item-weighted mean at 62.19. Gemini 3.1 Pro ranked second overall at 59.95, followed closely by Qwen 3.6 27B at 59.72. GPT-5.4 Mini, Gemini 3.5 Flash, and Qwen 3.5 397B-A17B obtained overall scores of 59.39, 58.86, and 57.04, respectively, whereas Grok 4.3 ranked lowest at 53.16.

GPT-5.4 ranked first in all three dimensions, scoring 70.42 in full-cycle cardiovascular decision-making and longitudinal care, 58.47 in multimodal interpretation and clinical documentation, and 43.17 in communication, humanistic care, safety, and ethics. Gemini 3.1 Pro ranked second in full-cycle care at 67.38 and second in multimodal and documentation tasks at 58.21 but sixth in communication, safety, and ethics at 37.41. Qwen 3.6 27B ranked third in full-cycle care at 67.10 and third in communication, safety, and ethics at 41.21. GPT-5.4 Mini showed the strongest relative specialization in the third dimension, ranking second at 42.47 despite placing fifth in full-cycle care.

Across all 7 LLMs, communication, humanistic care, safety, and ethics was the lowest-scoring dimension, with a cross-model mean of 39.70. Full-cycle care had the highest cross-model dimension mean at 65.68, followed by multimodal interpretation and clinical documentation at 55.88. The model ordering based on item-weighted means was identical to the macro-average ordering, indicating that task-size differences did not materially change the overall ranking.

\subsection{Task-level performance and between-model variability}

\begin{figure}
    \centering
    \includegraphics[width=1\linewidth]{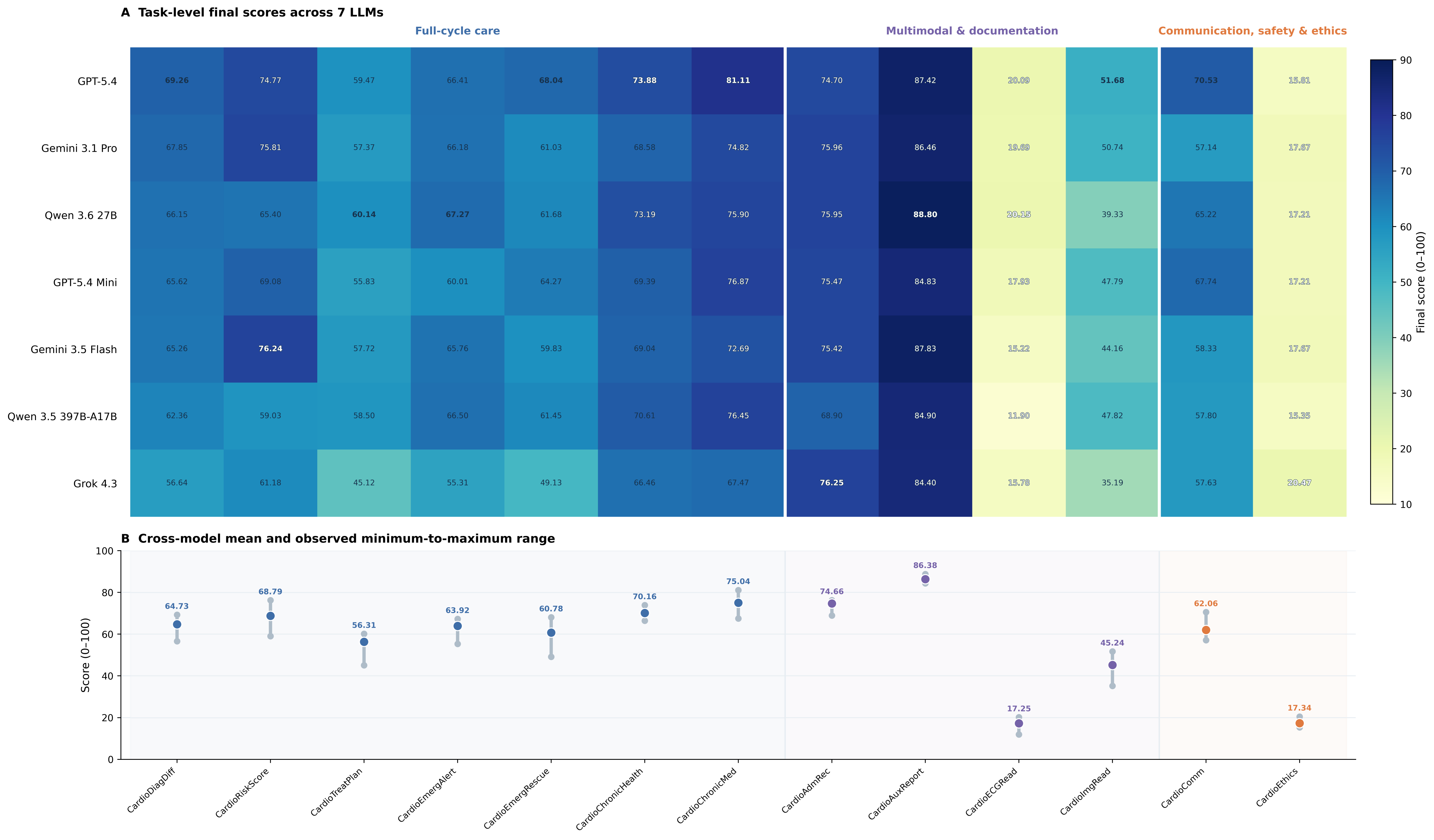}
    \caption{Task-level performance and between-LLM variability across CardioBench}
    \label{fig:figure3}
\end{figure}

Performance varied substantially across the 13 specialist tasks, with larger differences across tasks than across the overall LLM averages (Figure \ref{fig:figure3}). The highest cross-LLM mean was observed for CardioAuxReport at 86.38, followed by CardioChronicMed at 75.04, CardioAdmRec at 74.66, CardioChronicHealth at 70.16, and CardioRiskScore at 68.79. Figure \ref{fig:figure3}A shows that comparatively strong performance was concentrated in auxiliary report integration, admission documentation, risk scoring, and longitudinal management, although the leading LLM differed across these workloads.

The lowest cross-LLM means were observed for CardioECGRead at 17.25 and CardioEthics at 17.34. Cardiovascular image interpretation also remained challenging, with a mean score of 45.24. The consistently low ECG and ethics scores across the 7 LLMs indicate that these results were not attributable to a single poorly performing system. Instead, they suggest shared difficulty in specialized signal interpretation and context-sensitive ethical reasoning.

Between-LLM variability was not uniform across tasks (Figure \ref{fig:figure3}B). CardioEmergRescue showed the largest minimum-to-maximum range at 18.91 points, followed by CardioRiskScore at 17.21, CardioImgRead at 16.49, and CardioTreatPlan at 15.02. CardioChronicMed and CardioComm also showed notable ranges of 13.64 and 13.39 points, respectively. In contrast, CardioAuxReport had the highest mean score but a relatively narrow range of 4.40 points, indicating broadly strong performance across the evaluated LLMs.

The combined task-level visualization demonstrates that average task difficulty and between-LLM separation represent related but distinct properties. Some tasks, such as CardioAuxReport, were comparatively tractable for nearly all LLMs, whereas CardioRiskScore and CardioEmergRescue produced more pronounced differentiation. Conversely, CardioECGRead and CardioEthics combined low mean performance with limited between-LLM separation, suggesting broadly shared limitations rather than isolated system-specific failure.

\subsection{Divergence between holistic clinical quality and key-point coverage}

\begin{figure}
    \centering
    \includegraphics[width=1\linewidth]{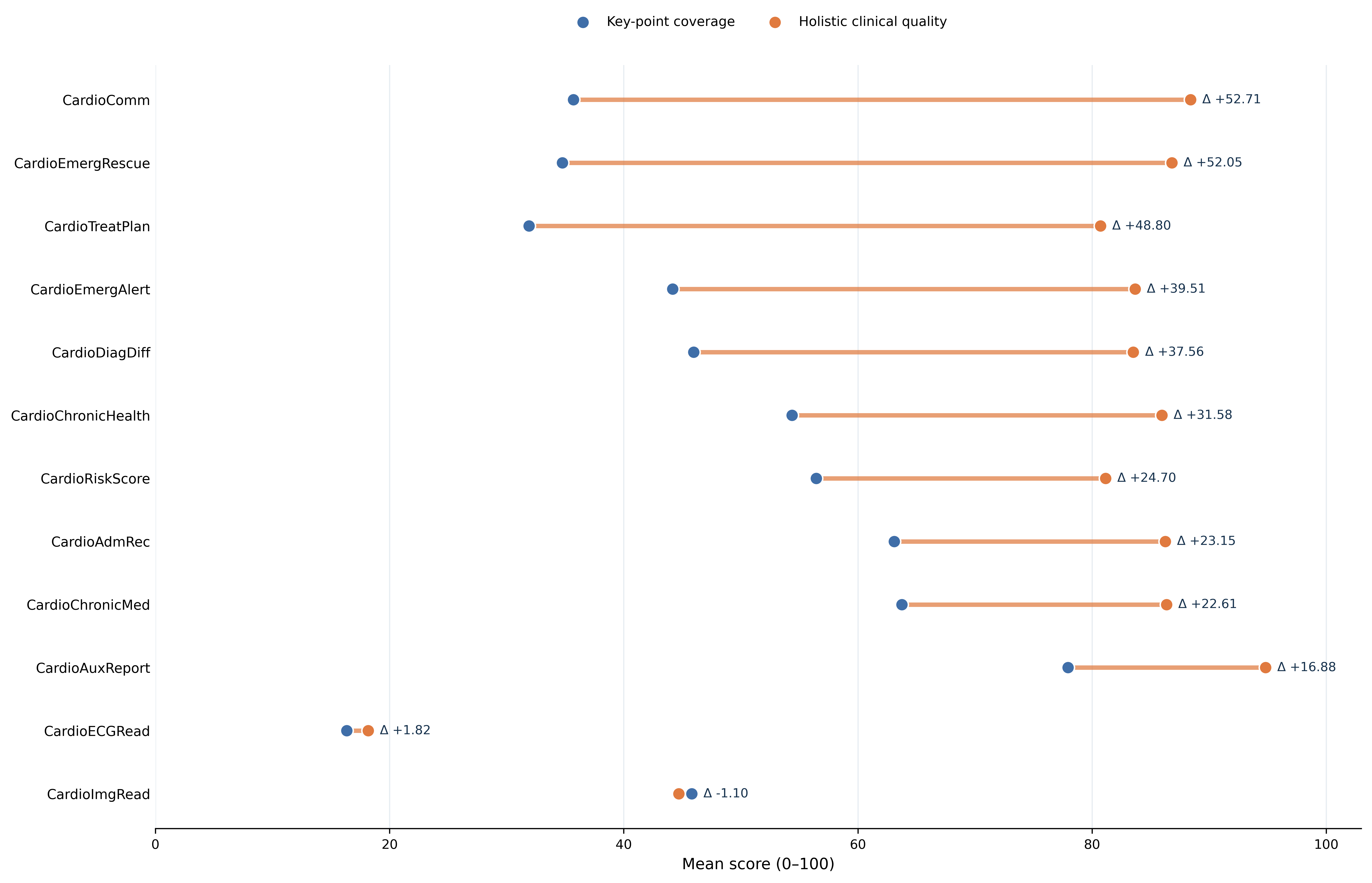}
    \caption{Divergence between holistic clinical quality and key-point coverage}
    \label{fig:figure4}
\end{figure}

The two components used to evaluate open-ended responses frequently produced different impressions of response quality (Figure \ref{fig:figure4}). The largest positive difference was observed for CardioComm, for which the mean holistic clinical quality score exceeded key-point coverage by 52.71 points. Similarly large gaps were identified for CardioEmergRescue at 52.05 points and CardioTreatPlan at 48.80 points.

Substantial differences were also observed for CardioEmergAlert at 39.51 points, CardioDiagDiff at 37.56 points, and CardioChronicHealth at 31.58 points. These tasks often require prioritization, integration of multiple clinical considerations, and organization of a coherent management sequence. Responses may therefore appear clinically plausible and well structured while failing to explicitly include all prespecified reference elements.

More moderate differences were found for CardioRiskScore, CardioAdmRec, and CardioChronicMed, with gaps of 24.70, 23.15, and 22.61 points, respectively. CardioAuxReport showed a smaller difference of 16.88 points. The two components were closely aligned for CardioECGRead, with a difference of 1.82 points, whereas CardioImgRead was the only task in which mean key-point coverage slightly exceeded holistic clinical quality, producing a difference of -1.10 points.

These findings indicate that holistic clinical quality and explicit key-point coverage assess different characteristics of generated responses. The former reflects overall clinical coherence, usability, and apparent plausibility, whereas the latter reflects whether specific expected clinical elements were explicitly included. Reporting both components therefore provides a more informative description of LLM behavior than relying on either measure alone.

\subsection{LLM-specific strengths and performance stability}

\begin{figure}
    \centering
    \includegraphics[width=1\linewidth]{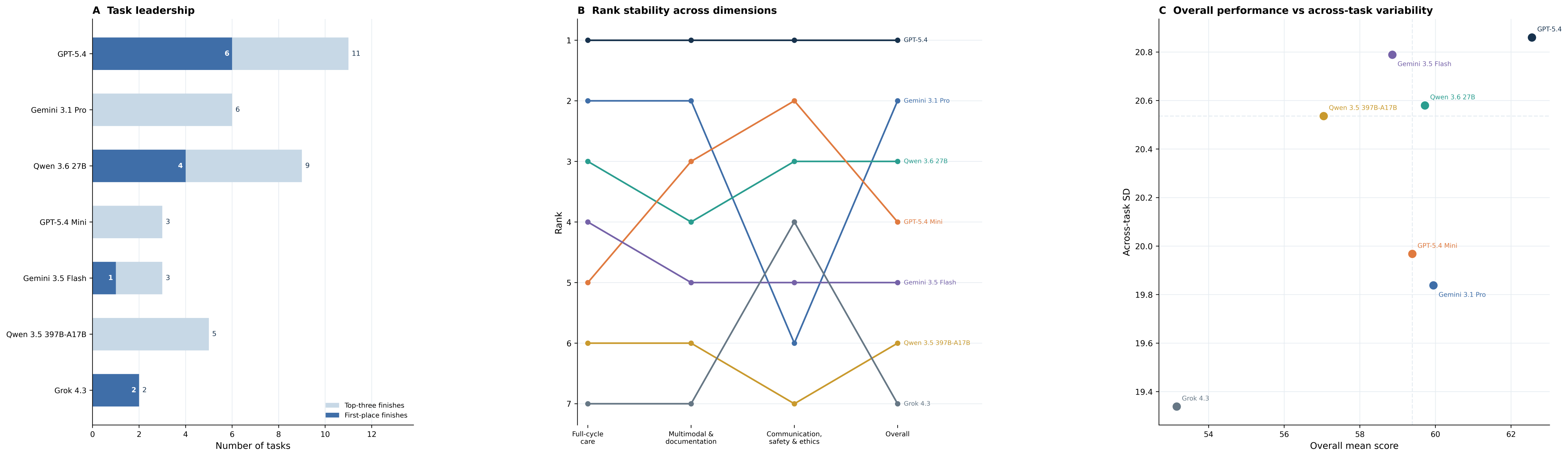}
    \caption{LLM-specific strengths and performance stability}
    \label{fig:figure5}
\end{figure}

No LLM achieved the highest score across all specialist workloads (Figure \ref{fig:figure5}A). GPT-5.4 ranked first in six tasks and placed among the top three in 11 of the 13 tasks, demonstrating the broadest task-level coverage. Qwen 3.6 27B led four tasks and achieved nine top-three placements. Grok 4.3 led two tasks, including admission record generation and clinical ethics, whereas Gemini 3.5 Flash achieved the highest score in cardiovascular risk scoring.

Several LLMs demonstrated competitive performance without frequently ranking first. Gemini 3.1 Pro did not lead an individual task but ranked among the top three in six tasks. Qwen 3.5 397B-A17B achieved five top-three placements, whereas GPT-5.4 Mini achieved three. These findings indicate that task wins alone do not fully capture the consistency or breadth of LLM performance.

LLM rankings also changed across clinical dimensions (Figure \ref{fig:figure5}B). GPT-5.4 ranked first in full-cycle care, multimodal interpretation and documentation, communication, safety and ethics, and the overall comparison. Gemini 3.1 Pro ranked second in the first two dimensions but sixth in communication, safety and ethics. Qwen 3.6 27B showed a more balanced pattern, ranking third, fourth, and third across the three dimensions and third overall. GPT-5.4 Mini improved from fifth in full-cycle care to third in multimodal and documentation tasks and second in communication, safety and ethics. Grok 4.3 ranked seventh overall but fourth in communication, safety and ethics, illustrating that lower overall rank did not preclude relative strength in a specific domain.

Across-task standard deviations ranged from 19.34 to 20.86, indicating substantial task-dependent variation for all 7 LLMs (Figure \ref{fig:figure5}C). GPT-5.4 combined the highest overall mean score of 62.55 with an across-task standard deviation of 20.86. Gemini 3.1 Pro and GPT-5.4 Mini had slightly lower variability, with standard deviations of 19.84 and 19.97, respectively. Grok 4.3 had both the lowest overall score and the lowest across-task standard deviation. Accordingly, lower variability should not be interpreted in isolation as evidence of stronger or more generalizable performance.

Overall, these analyses show that aggregate rankings conceal clinically meaningful differences in task specialization. The strongest LLM overall was not uniformly dominant, and several lower-ranked LLMs retained identifiable advantages in particular documentation, risk-scoring, communication, or ethics workloads.

\section{Discussion}
\subsection{Principal findings}

CardioBench provides a clinically structured evaluation of LLM performance across the cardiovascular care continuum. To our knowledge, it is the largest real-world, multi-task benchmark developed specifically for cardiology and provides the broadest coverage of clinically authentic cardiovascular scenarios reported to date. GPT-5.4 achieved the highest overall score and ranked first across all three dimensions, although its macro-average of 62.55/100 indicates that considerable opportunities for improvement remain. Performance varied substantially by task, with comparatively strong results in auxiliary report integration, chronic medication management, and admission documentation, whereas ECG interpretation and clinical ethics remained among the most challenging areas. Task-level leadership was distributed across several LLM families, and notable differences between holistic clinical quality and key-point coverage showed that fluent responses could still omit clinically important details. Although item-weighted and macro-averaged rankings were broadly consistent, dimension-level analyses revealed meaningful specialization.

The benchmark's contribution extends beyond scale. Existing cardiology evaluations have generally focused on examinations, individual modalities, guideline retrieval, isolated clinical decisions, or narrowly defined subspecialty cases\cite{skalidis2023chatgpt,sarraju2023opportunities,boonstra2024artificial,alexandrou2025performance,o2026large,zhang2025benchmarking}. CardioBench instead evaluates a linked sequence of documentation, diagnosis, risk assessment, treatment planning, deterioration recognition, emergency management, chronic care, multimodal interpretation, communication, and ethics within one framework. Its specialist-led annotation process further strengthens this design: 16 cardiology physicians with intermediate or higher professional titles performed task annotation and reference construction, while two cardiologists with senior professional titles conducted cross-review and adjudication. Separating primary annotation from senior verification reduced dependence on individual preferences and promoted consistent clinical standards across heterogeneous tasks.

\subsection{Documentation and structured integration were comparatively robust}

Auxiliary report integration was the strongest task, with all models scoring above 84 and a narrow cross-model range. Admission record generation also showed comparatively high performance. These tasks emphasize extraction, normalization, chronological organization, and synthesis of information that is explicitly present in the input. This pattern is consistent with evidence that adapted language models can perform strongly in clinical summarization\cite{van2024adapted}. It also suggests that documentation assistance may be a more realistic near-term use case than independent diagnostic or multimodal interpretation.

Nevertheless, documentation quality should not be equated with clinical safety. Even well-structured summaries may omit contradictions, uncertain findings, drug-related risks, or abnormal trends. Prospective deployment should therefore use source-linked drafting, clinician verification, and automatic checks for unsupported additions and missing critical values rather than treating a fluent note as a final record.

\subsection{ECG and cardiovascular image interpretation remain challenging}

ECG interpretation was the weakest task, with a cross-model mean of 17.25 and a maximum of only 20.15. The difference between the top two models was 0.06 points, indicating that the apparent task winner is not clinically meaningful. Raw ECG interpretation requires calibration awareness, spatial comparison across leads, rhythm analysis, interval measurement, morphology recognition, and integration of ischemic and conduction findings. The low results are consistent with recent evaluations concluding that generalist vision-enabled LLMs are not reliable substitutes for validated ECG algorithms, dedicated ECG systems, or expert interpretation\cite{engelstein2025effectiveness,soubh2026performance,lee2025comparative}.

Cardiovascular image interpretation performed better than ECG interpretation but remained moderate, with scores ranging from 35.19 to 51.68. This suggests that a model may extract some structured imaging information while remaining vulnerable to lesion localization, plaque characterization, stenosis grading, and image-quality limitations. Specialist multimodal development should combine validated image encoders, structured measurement tools, uncertainty estimation, and explicit reporting templates rather than relying on general-purpose visual prompting alone\cite{jin2024hidden,zhu2024multimodal,engelstein2025effectiveness,soubh2026performance}.

\subsection{Emergency tasks reveal a completeness problem rather than uniformly low plausibility}

The updated composite scores for early emergency recognition and emergency management were 63.92 and 60.78, respectively, suggesting that models could often produce clinically plausible responses. However, both tasks showed large judge-recall gaps, particularly emergency management, where the holistic judge mean was 86.80 but key-point recall was 34.75. This pattern indicates that models may identify the overall emergency and propose a reasonable plan while omitting time-critical details such as immediate stabilization steps, monitoring, contraindications, escalation thresholds, or sequence of actions.

Safety evaluation should therefore distinguish clinical plausibility from operational completeness. A response that sounds appropriate but omits cardiogenic shock recognition, defibrillation readiness, antithrombotic contraindications, or urgent reperfusion escalation may remain unsafe. Consequence-weighted benchmarks increasingly assign greater importance to high-risk omissions\cite{wang2025novel,asgari2025framework}. Future CardioBench versions could annotate each key point by urgency, potential harm, and recoverability and report a safety-weighted score alongside recall and holistic quality.

\subsection{Communication and ethics require separate validation}

Doctor-patient communication achieved a moderate composite mean of 62.06 and the largest judge-recall gap. Empathetic and coherent language may receive favorable holistic ratings even when informed consent elements, uncertainty, alternatives, warning signs, or follow-up responsibilities are incomplete. Healthcare-conversation frameworks recommend evaluating factual accuracy, comprehension, trust, empathy, personalization, and ethics as separate dimensions\cite{abbasian2024foundation}. CardioComm should therefore include both clinician review and patient-centered evaluation rather than relying on a single holistic score.

Clinical ethics remained a major outlier. Mean accuracy was 17.34\%, with values ranging from 15.35\% to 20.47\% on five-option questions. Only Grok 4.3 was marginally above the nominal 20\% random-choice reference. This pattern may reflect genuinely difficult scenarios, option ambiguity, keying or option-order problems, output normalization, or systematic reasoning failure. Aggregate results cannot distinguish these explanations. The ethics dataset requires item-level expert audit, answer-key verification, and formal comparison with chance before strong model-level conclusions are drawn. Cardiovascular professionalism and ethics guidance provides a specialty-relevant framework for this review\cite{benjamin20212020}.

\subsection{Measurement and deployment implications}

CardioBench demonstrates why specialist benchmarks should not collapse performance into one metric. Key-point recall is transparent and clinically auditable but sensitive to reference granularity and synonym matching. LLM-as-a-Judge can recognize semantically equivalent responses and global coherence but may overvalue verbosity or stylistic plausibility\cite{croxford2025evaluating,zhou2025automating,zheng2023judging}. The large task-specific gaps support reporting both components, conducting human calibration, and using task-specific rubrics. Communication tasks may require direct scoring of empathy and comprehension, whereas emergency tasks should emphasize critical actions and harmful omissions.

The benchmark does not support autonomous deployment of any evaluated model. Even the leading system showed severe limitations in ECG interpretation and ethics and incomplete coverage in emergency and communication tasks. The results are more compatible with selective augmentation under clinician oversight: structured report integration, draft documentation, checklist support, and retrieval of guideline-based information. Prospective evaluation should measure clinician-model team performance, time saved, error interception, automation bias, subgroup equity, and patient outcomes rather than model scores alone\cite{o2026large,johri2025evaluation,tam2024framework}.

CardioBench was developed using real-world data from a tertiary academic center and covers a broad range of cardiovascular tasks. Nevertheless, variations in patient populations, documentation practices, and clinical workflows across institutions may influence model performance, and future multicenter studies would further clarify the generalizability of the findings.

The comparison reflects the performance of 7 LLMs under standardized prompting and inference settings during the study period. As models and deployment strategies continue to evolve, subsequent evaluations may yield different results. The findings are therefore best interpreted as a controlled comparison within the defined study framework.

Open-ended responses were evaluated using complementary measures of clinical quality and expected-content coverage. This approach provides a structured assessment of both usability and completeness, although complex clinical responses may still benefit from additional expert review. Further validation with independent clinician panels and prospective clinical studies would provide additional evidence regarding real-world applicability.
\section{Conclusion}
CardioBench establishes a large-scale and clinically comprehensive framework for evaluating LLMs in cardiovascular care. To our knowledge, it is the largest real-world, multi-task cardiology benchmark spanning the cardiovascular care continuum and provides the broadest coverage of clinically authentic cardiology scenarios reported to date. Its 13 task-specific datasets capture the heterogeneous demands of documentation, diagnosis, risk assessment, treatment, emergency care, multimodal interpretation, longitudinal management, communication, and ethics. Annotation by 16 cardiology physicians with intermediate or higher professional titles, followed by cross-review from two cardiologists with senior professional titles, provides a rigorous foundation for objective, consistent, and reproducible evaluation. Across 2,263 items and 7 LLMs, performance remained highly task dependent, underscoring the value of task-specific validation, complementary metrics, expert review, and clinician oversight for responsible translation into cardiovascular workflows.
\section{Data Availability}
The CardioBench dataset is publicly available on the MedBench(\href{https://medbench.opencompass.org.cn/track-detail/12}{https://medbench.opencompass.org.cn/track-detail/12}) platform for model evaluation.

\bibliographystyle{unsrt}  
\bibliography{references}

\end{document}